\documentclass[letterpaper, 10 pt, conference]{ieeeconf}  
\usepackage{hhline}
\usepackage{graphicx}
\usepackage{subcaption}
\usepackage{multirow}
\usepackage{soul,color}
\usepackage{amsfonts}
\usepackage{float}
\usepackage{amsmath}
\usepackage{algorithm}
\usepackage{algorithmic}
\usepackage{boldline} 

\renewcommand\hl[1]{#1} 

\usepackage{kotex}

\usepackage{kotex-logo}

\IEEEoverridecommandlockouts                              

\title{\LARGE \bf
Transformer-based deep imitation learning for dual-arm robot manipulation
}
\author{Heecheol Kim$^{1}$$^{,}$$^{2}$, Yoshiyuki Ohmura$^{1}$, Yasuo Kuniyoshi$^{1}$%
\thanks{
This paper is based on results obtained under a Grant-in-Aid for Scientific Research (A) JP18H04108 of The University of Tokyo.
}
\thanks{$^{1}$ Laboratory for Intelligent Systems and Informatics, Graduate School of Information Science and Technology, The University of Tokyo, 7-3-1 Hongo, Bunkyo-ku, Tokyo, Japan (e-mail: \{h-kim, ohmura, kuniyosh\}@isi.imi.i.u-tokyo.ac.jp, Fax: +81-3-5841-6314) }%
\thanks{$^{2}$ Corresponding author}
}
\begin{document}
\maketitle
\thispagestyle{empty}
\pagestyle{empty}

\begin{abstract}
Deep imitation learning is promising for solving dexterous manipulation tasks because it does not require an environment model and pre-programmed robot behavior. However, its application to dual-arm manipulation tasks remains challenging.
In a dual-arm manipulation setup, the increased number of state dimensions caused by the additional robot manipulators causes distractions and results in poor performance of the neural networks. We address this issue using a self-attention mechanism that computes dependencies between elements in a sequential input and focuses on important elements. 
A Transformer, a variant of self-attention architecture, is applied to deep imitation learning to solve dual-arm manipulation tasks in the real world. 
The proposed method has been tested on dual-arm manipulation tasks using a real robot.
The experimental results demonstrated that the Transformer-based deep imitation learning architecture can attend to the important features among the sensory inputs, therefore reducing distractions and improving manipulation performance when compared with the baseline architecture without the self-attention mechanisms.
\end{abstract}

\providecommand{\keywords}[1]{\textbf{\textit{Index terms---}} #1}
 

\section{Introduction}
\begin{figure}
  \centering
  \vspace{0.0in}
  \includegraphics[width=0.98\linewidth]{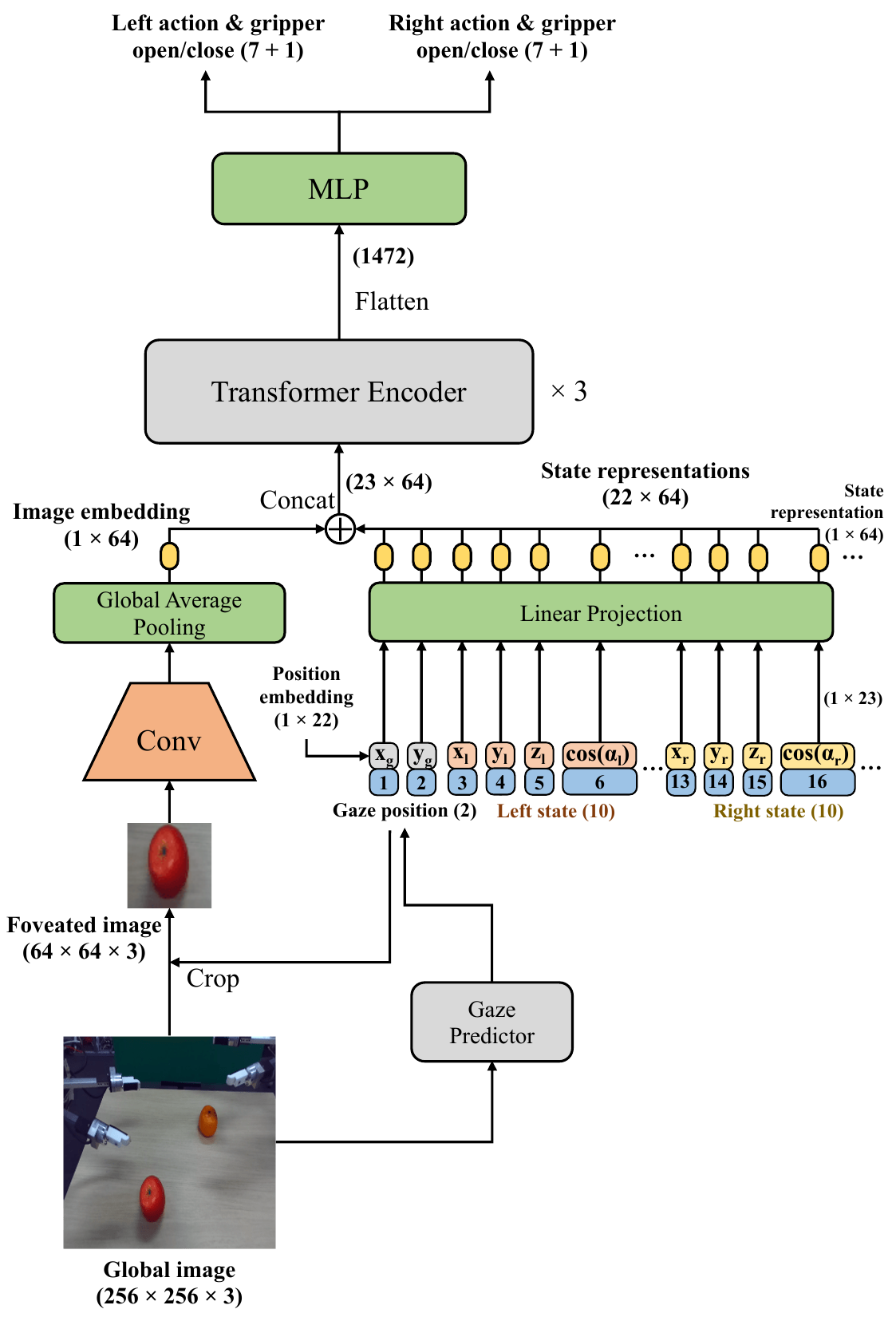}
  \captionsetup{justification=centering}
  \caption{Proposed Transformer-based deep imitation learning architecture for the dual-arm manipulation.}
  \label{fig:entire_architecture}
 \end{figure}

End-to-end deep imitation learning, which controls a robot by imitating an expert's demonstration, has gained popularity in the robotics community because of its potential application to dexterous manipulation tasks without the need to model environments or objects (e.g.,  \cite{yang2016repeatable,zhang2018deep,kim2021gaze}).

Dual-arm manipulation increases the manipulability of objects in many tasks \cite{smith2012dual} (e.g., peg-in-hole tasks \cite{suarez2015using}, cable deformation tasks \cite{Zhu2018dual}, and resolving occlusion with one hand while grasping an object with the other hand \cite{sepulveda2020robotic}). Because the human operator is familiar with dual-arm manipulation, it is easy to transfer the bimanual skills of a human using teleoperation \cite{smith2012dual}. Therefore, the imitation learning framework, which generates demonstration data by human teleoperation, is adequate for complex dual-arm manipulation tasks.
However, few studies have investigated the dual-arm manipulation framework using imitation learning (e.g., \cite{asfour2008imitation,caccavale2017imitation,xie2020deep}), and many of these focused on learning the hierarchical structure of the subtasks. However, we believe the problem to solve for deep imitation learning on dual-arm manipulation tasks lies in the distractions caused by the increased dimensions of the concatenated left/right robot arm kinematics states, which is essential for the collaboration of both arms for dual-arm manipulation. 

Our previous study revealed that the attention mechanism is important for the visuomotor control of robots with deep imitation learning \cite{kim2020using} to suppress distractions. This study measured the human gaze position with an eye-tracker while teleoperating the robot to generate demonstration data and used only the foveated vision around the predicted gaze position for deep imitation learning to suppress visual distraction from task-unrelated objects. 

\hl{
The gaze-based visual attention mechanism requires the robot's kinematics states from its somatosensory information because the peripheral vision should be masked out to suppress visual distractions. However, in the dual-arm deep imitation learning framework, distraction is also caused by high dimensional kinematics states. For example, when the robot reaches its right arm to an object, the left arm kinematics states are not used to compute the policy and become distractions. }
\hl{Unlike visual attention which can be trained with gaze, there is no such teaching signal for somatosensory information. Thus, a novel method to learn somatosensory attention in an unsupervised way is required.}

Self-attention architectures, especially the Transformer \cite{vaswani2017attention}, evaluate relationships between elements on an input sequence. 
Self-attention was first introduced to solve natural language processing (e.g., \cite{devlin2018bert}, \cite{brown2020language}, \cite{dosovitskiy2020image}), and has since been widely adopted in many different domains such as image classification \cite{dosovitskiy2020image}, object detection \cite{carion2020end}, and high-resolution synthesis \cite{esser2020taming}, and learning multiple domains with one model \cite{kaiser2017one}. This wide application of the self-attention to different domains may imply that it can also be used to suppress distractions in kinematics states, which is a problem in dual-arm manipulation.

To this end, we propose a Transformer-based attention mechanism for sensory information for dual-arm robot imitation learning (Fig. \ref{fig:entire_architecture}). 
\hl{Because the transformer dynamically generates attention based on the input state, this architecture can be applied to suppress distractions on kinematics states without attention signal.}
In this architecture, each element of the sensory inputs (gaze position, left arm state, and right arm state) as well as the foveated image embedding is input to the Transformer to determine which element should be paid attention. This attention makes the output policy robust to unnecessary distractions in the sensory inputs.

The proposed Transformer-based deep imitation learning architecture was evaluated on an uncoordinated manipulation task (two arms executing different tasks), a goal-coordinated manipulation task (both arms solving the same task but not physically interacting with each other), 
and bimanual tasks (both arms physically interacting to solve the task) \cite{smith2012dual}. The results demonstrate that the Transformer-based self-attention mechanism improves dual-arm manipulation.

\section{Related work}
\subsection{Imitation learning-based dual-arm manipulation}

Previous imitation learning approaches to dual-arm manipulation mainly focused on the acquisition of subtask structures. Reference \cite{asfour2008imitation} studied the imitation learning of dual-arm manipulation tasks based on keypoints selected by hidden Markov models (HMM) to reproduce the movements of arms demonstrated by the human demonstrator. An imitation learning framework was proposed by \cite{caccavale2017imitation} that segments motion primitives and learns task structure from segments for dual-arm structured tasks in simulation. The proposed framework was tested on a pizza preparation scenario, which is an uncoordinated manipulation task. Finally, \cite{xie2020deep} designed a deep imitation learning model that captures relational information in dual-arm manipulation tasks to improve bimanual manipulation tasks in the simulated environment, but their work required manually defined task primitives.
To the best of our knowledge, the performance of a self-attention-based deep imitation learning method for dual-arm manipulation has not been studied in a real-world robot environment.

\subsection{Transformer-based robot learning}
The Transformer architecture has been used in robot learning for vision
\cite{dasari2020transformers} in a simulated robot environment to bridge the domain gaps, such as different morphologies, physical appearances, or locations, between the demonstrator and the target robot.
In addition, \cite{Cachet2020Transformer} used the Transformer-based seq-to-seq architecture to improve meta-imitation learning. Here, they used the Transformer to capture temporal correspondences between the demonstration and the target task. 

\hl{
However, these studies did not apply the self-attention architecture to robot kinematics states.
The robot kinematics states provide essential information which cannot be captured by a gaze-based visual attention system, which reduces visual distraction by masking out peripheral vision. Therefore, our work aims to design a Transformer-based architecture to determine the attention to multiple sensory inputs in the current state for robust output against the distractions caused by the increased number of robot kinematics states.}

\section{Method}

\subsection{Robot system}
We use a dual-arm robot system designed for teleoperation and imitation learning \cite{kim2021gaze,kim2020using}. While a human operator teleoperates two UR5 (Universal Robots) manipulators, a head-mounted display (HMD) provides vision captured from a stereo camera mounted on the robot. During teleoperation, the human gaze is measured by an eye-tracker mounted in the HMD.
In this research, the left camera image is resized into $256 \times 256$ (called the global image) and recorded at 10 Hz with the two-dimensional gaze coordinate of the left eye and robot kinematics states of both arms. Each robot kinematics state is defined as a ten-dimensional vector that represents the end-effector position (three dimensions), orientation (six dimensions, represented using a combination of cosines and sines of three-dimensional Euler angles to prevent drastic changes of states when the angle exceeds $2 \pi$), and the gripper angle (one dimension) for each arm. 

\subsection{Gaze position prediction} \label{section:gaze}
In \cite{kim2020using}, human gaze was used to achieve imitation learning of a robot manipulator that is robust against visual distractions. This approach first predicts gaze position and crops the images around the predicted gaze position to remove task-irrelevant visual distractions, such as objects unseen during training. It was proved that such a gaze mechanism can improve the manipulation of multi-object tasks.

Like \cite{kim2020using}, the current study also uses a mixture density network (MDN) \cite{bishop1994mixture}, which is a neural network architecture that fits a Gaussian mixture model (GMM) into the target, for estimating the probability distribution of the gaze position. The gaze predictor (Fig. \ref{fig:gaze}) \hl{inputs the entire $256 \times 256 \times 3$ RGB image and outputs $\mu \in \mathbf{R}^{2\times N}, \sigma \in \mathbf{R}^{2 \times N}, \rho \in \mathbf{R}^{1 \times N}, p \in \mathbf{R}^N$}, \hl{which comprises the probability distribution of a two-dimensional gaze coordinate location where $N$ is the number of Gaussian distributions that compose the GMM. We used $N=8$ throughout this research.}
Because the gripper state may be informative for gaze prediction on subtask transitions (e.g., the gaze shifts to a toy orange as soon as the left gripper has closed to pick up a toy apple, on \textit{Pick} task), gripper angles are added. 
This network is trained by minimizing the negative log-likelihood of the probability distribution with the measured human gaze as target $e$ \hl{as follows}:

\begin{equation}
\label{eq:loss_gaze}
\mathcal{L}_{gaze} = -log \big{(}\sum_{i=1}^N p^i \mathcal{N} (e;\mu^i,\mathbf{\Sigma}^i)\big{),}
\end{equation}
\hl{where} $\mathbf{\Sigma}^i=\begin{bmatrix} {\sigma^i_x}^2 & \rho^i \sigma^i_x \sigma^i_y \\ \rho^i \sigma^i_x \sigma^i_y & {\sigma^i_y}^2 \end{bmatrix}$ represents the covariance matrix.

During the test, a single gaze position is selected from the \hl{mean} location \hl{($\mu$)} with the highest probability of the inferred \hl{GMM}:
\begin{align*}
\label{eq:gaze_selection}
{gaze} = \mu^{{arg\,max} (p^i)}
\end{align*}

\subsection{Transformer-based dual-arm imitation learning}

\begin{figure}
  \centering
  \vspace{0.0in}
  \begin{subfigure}[t]{0.25\textwidth}
    \captionsetup{width=.8\linewidth}
    \captionsetup{justification=centering}
    \includegraphics[width=0.98\linewidth]{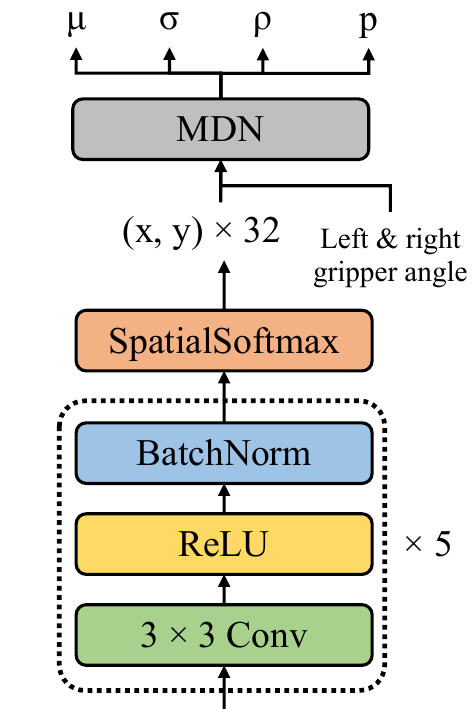}
    \caption{Gaze predictor}
    \label{fig:gaze}
  \end{subfigure}%
  \begin{subfigure}[t]{0.25\textwidth}
    \captionsetup{width=.8\linewidth}
    \captionsetup{justification=centering}
    \includegraphics[width=0.98\linewidth]{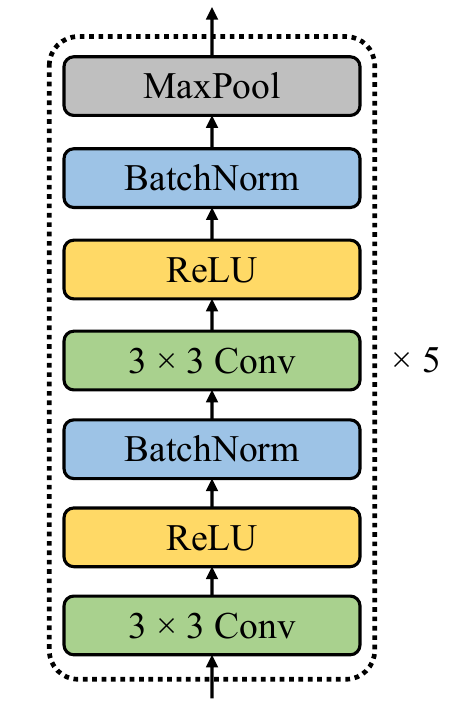}
    \caption{Convolution block}
    \label{fig:conv}
  \end{subfigure}%

  \captionsetup{justification=centering}
  \caption{Neural network architectures.}
 \end{figure}
 
The proposed architecture for dual-arm manipulation is presented in Fig. \ref{fig:entire_architecture}. First, the gaze predictor (\ref{section:gaze}) predicts two-dimensional gaze position from the entire $256 \times 256 $ image. The $64 \times 64$ foveated image, cropped according to the predicted gaze position, passes through five convolution blocks (Fig. \ref{fig:conv}) with channel sizes of $[8, 16, 16, 32, 64]$ in that order. The resultant feature passes through the global average pooling (GAP) layer \cite{lin2013network}, which averages each feature map into one value to reduce the number of parameters.

The predicted gaze position and left/right robot manipulator states are concatenated into a 22-dimensional state \hl{(Fig. }\ref{fig:entire_architecture}).
\hl{In the state representation, $[$$x_{gaze}$, $y_{gaze}$$]$ indicates gaze position, $[$$x_l$, $y_l$, $z_l$, $cos(\alpha_l)$, $sin(\alpha_l)$, $cos(\beta_l)$, $sin(\beta_l)$, $\cos(\gamma_l)$, $\sin(\gamma_l)$, $g_l$$]$ indicates left robot state and $[$$x_r$, $y_r$, $z_r$, $cos(\alpha_r)$, $sin(\alpha_r)$, $cos(\beta_r)$, $sin(\beta_r)$, $\cos(\gamma_r)$, $\sin(\gamma_r)$, $g_r$$]$ indicates right robot state, where $\alpha$, $\beta$, $\gamma$ represents the Euler angle and $g$ indicates gripper angle.}

In previous researches, the Transformer architecture was used to learn dependencies between word embeddings or embeddings of image patches \cite{vaswani2017attention,dosovitskiy2020image}. 
In this work, to learn dependencies between each element of the robot state, robot states were separated into scalar values. \hl{And we added a 22-dimensional one-hot vector as positional embedding to each value.}
Each \hl{(1+22)}-dimensional state element with a one-hot vector that represents position passes through a linear projection to generate a 64-dimensional state representation with the learned position embeddings. Then, the image embedding and state representations are concatenated and encoded using the three layers of the Transformer encoder \hl{(see Fig }\ref{fig:entire_architecture}). We followed the Transformer encoder architecture described in \cite{vaswani2017attention}.

Finally, the encoded feature is flattened and passes through an MLP with one hidden \hl{layer} to predict the actions for both arms. An action is defined as the difference between the next kinematics state and the current kinematics state: $a_t = s_{t+1} - s_t$, where $s_t$ represents the state vector at time step $t$ and $a_t$ represents the action at time step $t$. Unlike the angles in the state representation, which are represented by a combination of $cos(\theta)$ and $sin(\theta)$, where $\theta$ is an Euler angle, the angles on the action are represented by the differences in a three-dimensional Euler angle.

The gripper is controlled by the last element of the predicted action. However, this element only predicts the angle of the gripper and does not provide enough force to grasp any object. Therefore, a binary signal for the gripper open/close command is also predicted. If this binary signal predicts that the gripper should be closed, the gripper command additionally tries to close to $5^\circ$ to provide enough force to grasp objects. 

We used the same loss function first proposed in  \cite{zhang2018deep} and used in \cite{kim2021gaze}, \cite{kim2020using} to optimize the policy network.

\section{Experiments}
\subsection{Task setup}
We selected dual-arm manipulation tasks that include uncoordinated, goal-coordinated, and bimanual dual-arm manipulation tasks.

\begin{table}
\centering
\begin{tabular}{lll}
\hlineB{2}
    Dataset                   & Number of demos & Total demo time (min) \\ \hline \hline
\textit{Pick}                  & 1,030     & 85.4                  \\
\textit{BoxPush} & 460     & 53.8                   \\ 
\textit{ChangeHands} & 256     & 25.1                   \\ 
\hline \hline
\textit{KnotTying} & & 262.8 in total \\
\multicolumn{1}{r}{\textit{PickUp}}   & 1,858     & 77.1    \\
\multicolumn{1}{r}{\textit{Grasp}} & 2,251     & 98.9      \\ 
\multicolumn{1}{r}{\textit{PullOut}} & 607     & 16.5    \\ 
\multicolumn{1}{r}{\textit{ReleaseAndPick}} & 1,950     & 70.3         \\ 
\hlineB{2}

\end{tabular}
\caption{Training set statistics. 
To reduce the preparation time required for \textit{KnotTying}, the task was segmented into subtasks (\textit{PickUp}, \textit{Grasp}, \textit{PullOut}, and \textit{ReleaseAndPick}), and demonstrations were separately recorded for each subtask.
}
\label{tab:dataset_statistics}
\end{table}
The recorded demonstrations were divided into training set (90\%) and validation set (10\%).
The training set statistics of each task are represented in Table \ref{tab:dataset_statistics}. 
During the test, the target object was located as close as possible to the initial position of the target recorded on the randomly shuffled validation set images. For \textit{KnotTying}, the initial position images were manually chosen to avoid too complicated knot placements.

\hl{The size of the MLP hidden layer is 200 for \textit{Pick}, \textit{BoxPush} and \textit{ChangeHands}, and 400 for \textit{KnotTying} because \textit{KnotTying} requires a more complicated policy.} The models were trained with distributed training supported by NVIDIA Apex with a Xeon CPU E5-2698 v4 and eight V100 GPUs, using a batch size of 64 per GPU. During testing on the real robot system, Intel CPU Core i7-8700K and one NVIDIA GeForce GTX 1080 Ti were used. We used a learning rate of $10^{-5}$ and the rectified Adam (RAdam) optimizer \cite{liu2019variance}.

\subsection{Baselines}
We compared the proposed model with two baseline models without the Transformer encoder.
\hl{The first baseline model (baseline-GAP) retains the GAP layer but replaces each Transformer encoder layer with the one fully connected layer. Therefore, in this model, the foveated image is processed with a convolution block; averaged by the GAP layer; concatenated with the gaze position, the left state, and the right state; and processed by five fully connected layers to output the action. The second baseline model (baseline) does not include GAP for image processing nor the Transformer encoder. In this model the foveated image is processed with a convolution block; flattened and concatenated with the gaze position, the left state, and the right state; and then processed with the MLP with one hidden layer to finally compute the action output. }


The sizes of the hidden layer of both models were adjusted so that the total number of parameters is similar to the proposed Transformer-based architecture.

\subsection{Performance evaluation}
 
\begin{figure*}
\centering 
\vspace{0.1in}
\includegraphics[width=1.0\linewidth]{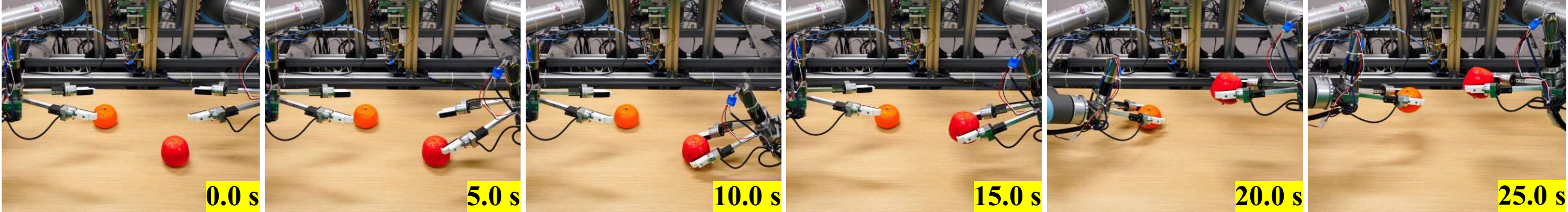}
\caption{Example of the proposed method on \textit{Pick}. The robot first picked the toy apple ($\sim10.0$s), lifted it up ($15.0$s), and picked up the orange ($25.0$s).}
\label{fig:apple_orange}
\end{figure*}

\begin{figure*}
\centering 
\vspace{0.1in}
\includegraphics[width=1.0\linewidth]{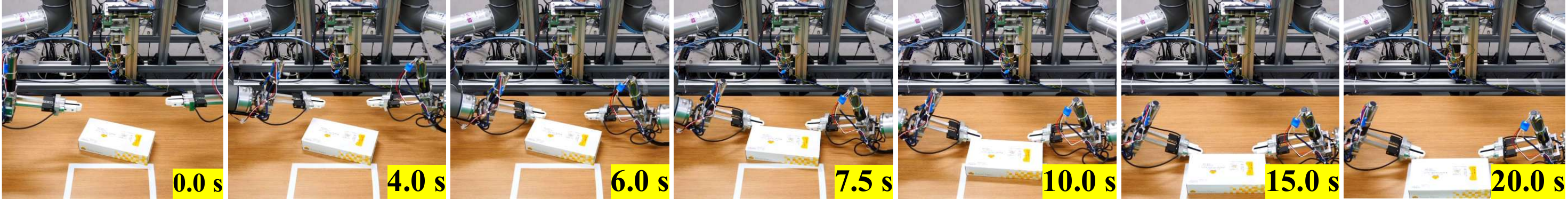}
\caption{Example of the proposed method on \textit{BoxPush}. The robot placed its both arms behind the box ($\sim6.0$s), pushed it with the right arm ($7.5$s), and moved it with both arms to the goal position ($\sim20.0$s).}
\label{fig:box_push}
\end{figure*}

\begin{figure*}
\centering 
\vspace{0.1in}
\includegraphics[width=1.0\linewidth]{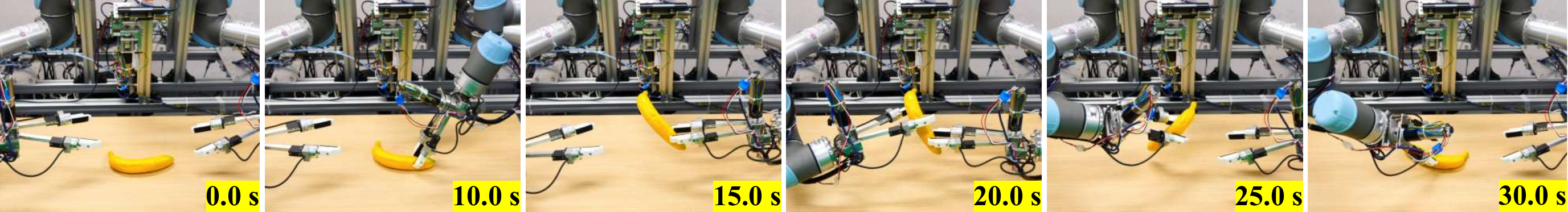}
\caption{Example of the proposed method on \textit{ChangeHands}. The robot first grasped the toy banana with its left hand ($\sim10.0$s), standed it up ($15.0$s), regrasped it with the right hand ($20.0$s $\sim$ $25.0$s), and finally flipped it ($30$s).}
\label{fig:banana_flip}
\end{figure*}

\begin{figure*}
\centering 
\vspace{0.1in}
\includegraphics[width=1.0\linewidth]{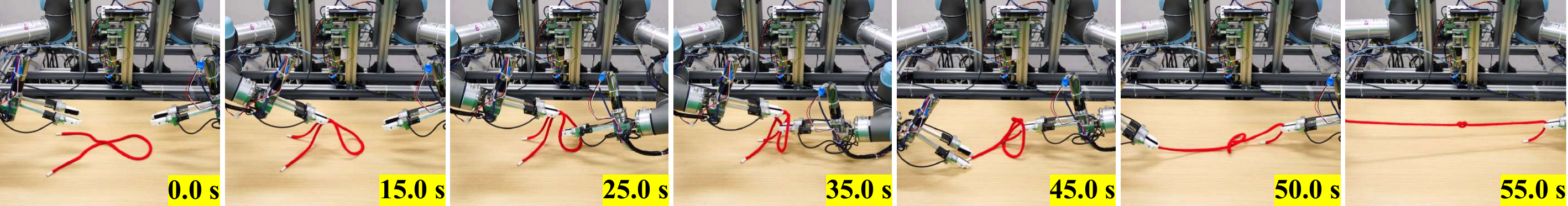}
\caption{Example of the proposed method on \textit{KnotTying}. The robot picked up the knot with its right arm ($\sim15.0$s), tried to grasp the end of the knot with left arm ($25.0$s), pull it ($35.0$s), release the right gripper and grasp the the other end of the knot ($\sim50.0$s), and finally tie it by stretching both arms ($55.0$s, programmed behavior).}
\label{fig:knot_tying}
\end{figure*}

\begin{figure*}
  \centering
  \vspace{0.0in}

  \begin{subfigure}[t]{0.45\linewidth}
    \captionsetup{width=.8\linewidth}
    \captionsetup{justification=centering}
    \includegraphics[width=0.98\linewidth]{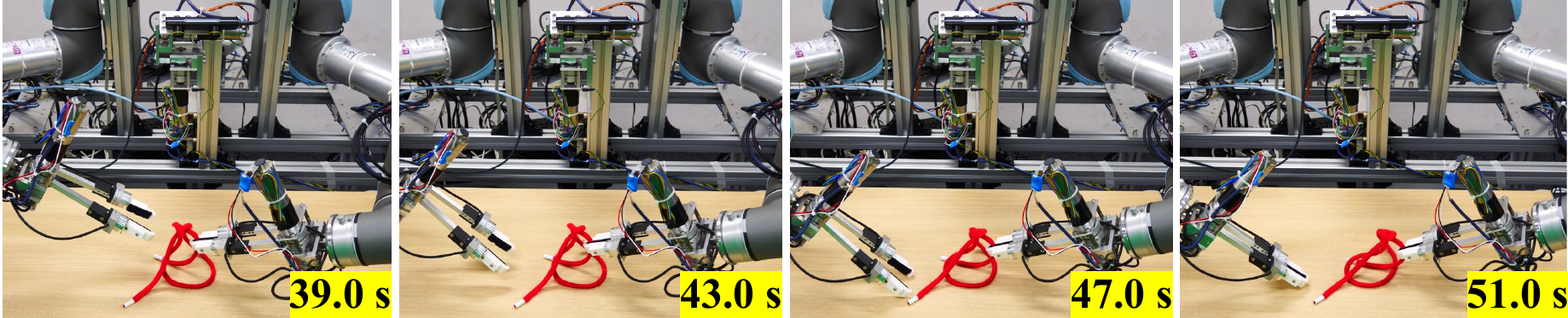}
    \caption{\hl{Failure at \textit{ReleaseAndPick}}}
    \label{fig:knot_tying_failure1}
  \end{subfigure}%
  \begin{subfigure}[t]{0.45\linewidth}
    \captionsetup{width=.8\linewidth}
    \captionsetup{justification=centering}
    \includegraphics[width=0.98\linewidth]{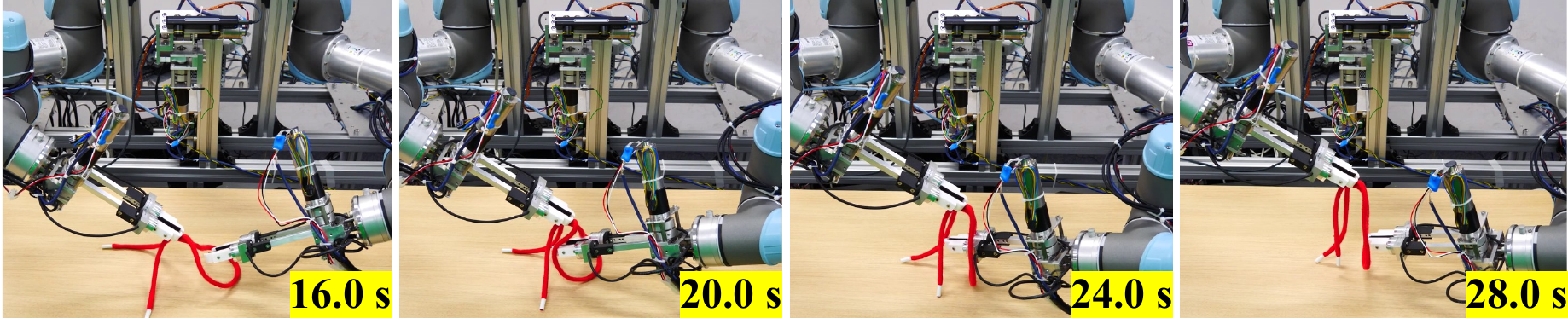}
    \caption{\hl{Failure at \textit{Grasp}}}
    \label{fig:knot_tying_failure0}
  \end{subfigure}%

  \captionsetup{justification=centering}
  \caption{\hl{Failure examples of the proposed method on \textit{KnotTying}. Typical failure was caused by mistakes on reaching to a target location of the knot.}}
  \label{fig:knot_tying_failure}
 \end{figure*}

\begin{figure}
  \centering
  \vspace{0.0in}
  \includegraphics[width=0.98\linewidth]{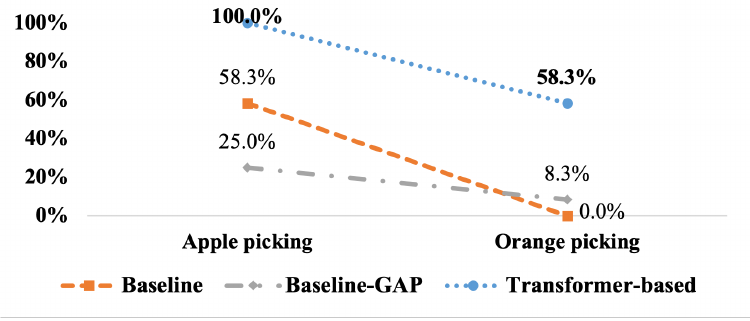}
  \captionsetup{justification=centering}
  \caption{Success rate on \textit{Pick} (24 trials).}
  \label{fig:apple_orange_result}
 \end{figure}

\begin{table}
\centering
\begin{tabular}{lll}
\hlineB{2}
Metric                      & Baseline & Transformer-based \\ \hline \hline
Orientation error $(^\circ)$                       & 9.57     & \textbf{1.99}  \\            
Top-left position error (mm)  & 35.7     & \textbf{21.3}      \\        
Top-rihgt position error (mm) & 36.4     & \textbf{12.2}           \\ \hlineB{2}
\end{tabular}
\caption{\textit{BoxPush} evaluation results (median of 24 trials).}
\label{tab:box_push}
\end{table}

\begin{figure}
  \centering
  \vspace{0.0in}
  \includegraphics[width=0.98\linewidth]{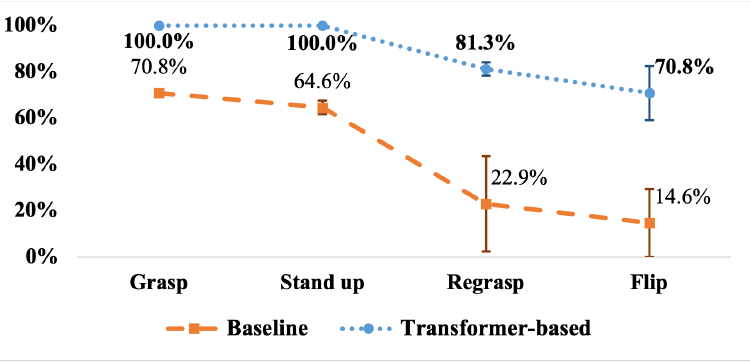}
  \captionsetup{justification=centering}
  \caption{Success rate on \textit{ChangeHands} (24 trials).}
  \label{fig:banana_flip_res}
 \end{figure}

\begin{figure}
  \centering
  \vspace{0.0in}
  \includegraphics[width=0.98\linewidth]{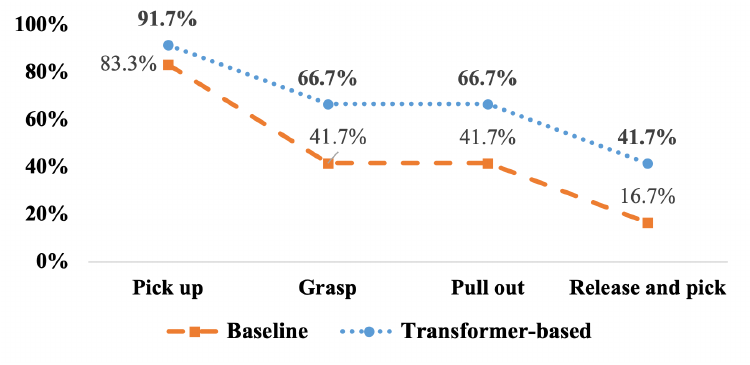}
  \captionsetup{justification=centering}
  \caption{Success rate on \textit{KnotTying} (12 trials).}
  \label{fig:knot_tying_result}
 \end{figure}

The performance of the Transformer-based method and baselines were evaluated for all tasks. Each method was tested on 24 different initial positions sampled from the validation set for \textit{Pick}, \textit{BoxPush}, \textit{ChangeHands} and 12 different initial positions for \textit{KnotTying}. In task \textit{Pick}, the Transformer-based method, baseline, and baseline-GAP models attempted to first pick up the toy apple with the left arm and then pick up the toy orange with the right arm (Fig. \ref{fig:apple_orange}). The result indicates that the Transformer-based method outperforms both baseline methods in terms of success rates of picking up each object (Fig. \ref{fig:apple_orange_result}).
Because the baseline-GAP performed the worst among the three models, it was excluded from further experiments. 

In \textit{BoxPush}, the Transformer-based method and the baseline were compared (Fig. \ref{fig:box_push}). For quantitative evaluation, the final positions of the top-left corner and top-right corner of the box after the robot executed the task were measured and the Euclidean distance to the ideal target position of both corners was calculated. Then, the orientation error, which measures how much the box is tilted with respect to the ideal target orientation (approximately $0^\circ$), was calculated based on the two measured positions. The result (Table. \ref{tab:box_push}) indicates that the Transformer-based method outperforms the baseline.

In \textit{ChangeHands}, the success rates of each subtask: \textit{Grasp} (left hand), \textit{StandUp} (left hand), \textit{Regrasp} (right hand), and \textit{Flip} (right hand), were evaluated (Fig. \ref{fig:banana_flip}), with models trained with two different random seeds (0, 1). Fig. \ref{fig:banana_flip_res} shows that the Transformer-based method achieved a high success rate. 
The result also confirms that the Transformer-based model outperforms the baseline regardless of the random seed.

In \textit{KnotTying}, the success rates of subtasks (\textit{PickUp}, \textit{Grasp}, \textit{PullOut}, and \textit{ReleaseAndPick}) were evalulated (Fig. \ref{fig:knot_tying}). The Transformer-based method recorded higher success rate than the baseline (Fig. \ref{fig:knot_tying_result}). \hl{Fig. }\ref{fig:knot_tying_failure} \hl{describes typical mistakes of the proposed method on \textit{KnotTying}. Among the total of 12 trials, we observed 1 failure at \textit{PickUp}, 3 failures at \textit{Grasp}, and 3 failures at \textit{ReleaseAndPick}.}

We have found that imitating a releasing behavior during \textit{ReleaseAndPick} was not successful, probably because the exact release timing is not strongly related to current sensory input. Rather, the human operator can release the gripper anytime after the subtask \textit{PullOut} is completed, causing the release signal to be diluted. Because our purpose is to adapt self-attention to current sensory inputs, we programmed the release behavior defined by opening the right hand for only one time if both hands are closed and the right hand is on the right side of the left hand with some margin.

\subsection{Attention weight assessment}
\begin{figure}
  \centering
  \includegraphics[width=0.98\linewidth]{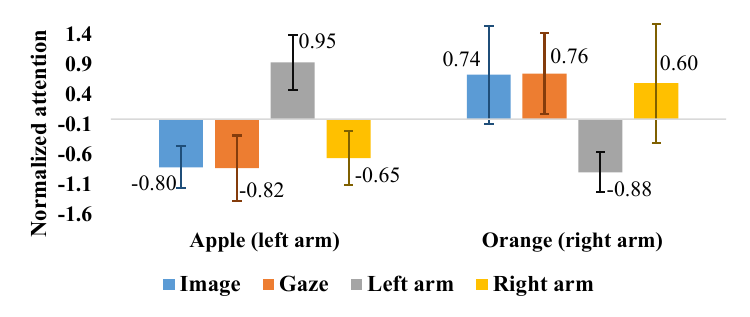}
  \captionsetup{justification=centering}
  \caption{\hl{Attention} for the subtasks in \textit{Pick}.}
  \label{fig:apple_orange_attention_weights}
 \end{figure}

\begin{figure}
  \centering
  \includegraphics[width=0.98\linewidth]{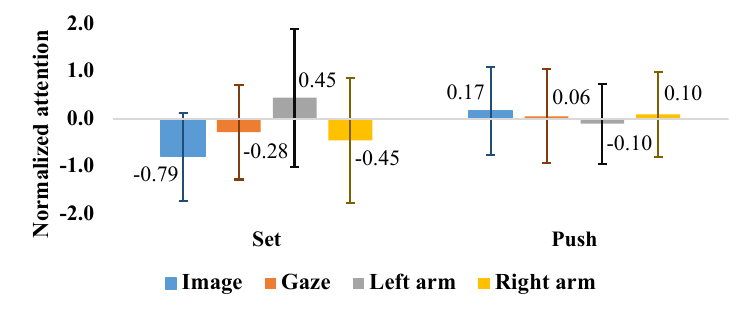}
  \captionsetup{justification=centering}
  \caption{\hl{Attention} for the subtasks in \textit{BoxPush}.}
  \label{fig:box_attention_weights}
 \end{figure}

\begin{figure}
  \centering
  \includegraphics[width=0.98\linewidth]{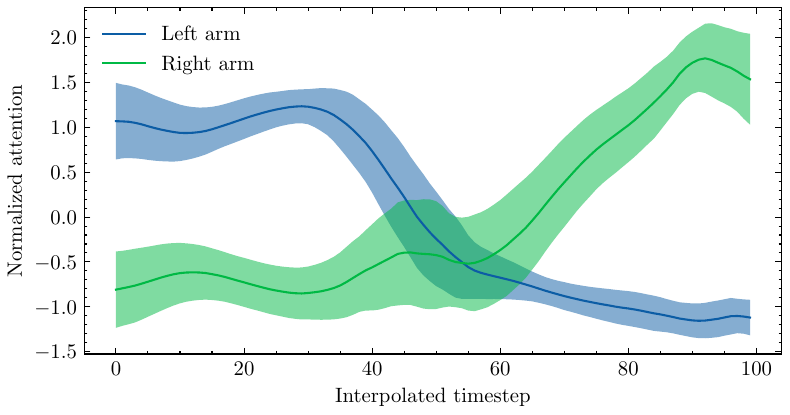}
  \captionsetup{justification=centering}
  \caption{\hl{Time series of attention for \textit{Pick}.}}
  \label{fig:apple_orange_temporal_attention}
 \end{figure}

To determine which sensory input the Transformer attends, we investigated the attention weights for each sensory input.
First, attention rollout \cite{abnar2020quantifying}, which is a recursive multiplication of the attention weights of all Transformer layers averaged over all attention heads, was computed. \hl{Next, the attention value for sensory inputs belong to the foveated image, gaze position, left arm state, and right arm state was calculated from $23 \times 23$ attention rollout as follows:} 
\begin{equation}
\begin{aligned}
\label{eq:summed}
&W^{Image} = \sum^{23}_{i=1} A_{i1}\\
\end{aligned}
\end{equation}
\begin{equation}
\begin{aligned}
\label{eq:summed}
&W^{Gaze} = \sum^{23}_{i=1} \sum^{3}_{j=2} A_{ij}\\
\end{aligned}
\end{equation}

\begin{equation}
\begin{aligned}
\label{eq:summed}
W^{Left} = \sum^{23}_{i=1}\sum^{13}_{j=4} A_{ij}\\
\end{aligned}
\end{equation}

\begin{equation}
\begin{aligned}
\label{eq:summed}
W^{Right} = \sum^{23}_{i=1} \sum^{23}_{j=14} A_{ij},\\
\end{aligned}
\end{equation}
where $A$ represents the attention rollout map in which the columns represent queries and the rows represent values, and $W$ refers to \hl{attention value for each input domain}.
\hl{Then, each time series attention values in each input domain (\textit{Image}, \textit{Gaze}, \textit{Left}, \textit{Right}) was normalized because we want to see the change of attention values on each input domain:}
\begin{equation}
\begin{aligned}
\label{eq:normed}
W'_t = \frac{W_t - \mu_W}{\sigma_W},\\
\end{aligned}
\end{equation}
\hl{for $t \in [0, 1, ..., T]$ in each trial episode. $\mu_W$ and $\sigma_W$ refers mean and standard deviation of time series attention values $[W_0, W_1, ..., W_T]$.}

 \begin{figure}
  \centering
  \vspace{0.0in}
  \begin{subfigure}[t]{0.20\textwidth}
    \captionsetup{width=.8\linewidth}
    \captionsetup{justification=centering}
    \includegraphics[width=0.98\linewidth]{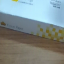}
    \caption{\textit{Set}: foveated image}
    \label{fig:set_foveated}
  \end{subfigure}%
  \begin{subfigure}[t]{0.20\textwidth}
    \captionsetup{width=.8\linewidth}
    \captionsetup{justification=centering}
    \includegraphics[width=0.98\linewidth]{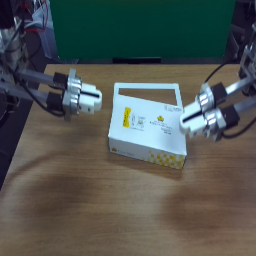}
    \caption{\textit{Set}: global image}
    \label{fig:set_camera}
  \end{subfigure}%
  \vskip\baselineskip

  \begin{subfigure}[t]{0.20\textwidth}
    \captionsetup{width=.8\linewidth}
    \captionsetup{justification=centering}
    \includegraphics[width=0.98\linewidth]{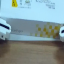}
    \caption{\textit{Push}: foveated image}
    \label{fig:push_foveated}
  \end{subfigure}%
  \begin{subfigure}[t]{0.20\textwidth}
    \captionsetup{width=.8\linewidth}
    \captionsetup{justification=centering}
    \includegraphics[width=0.98\linewidth]{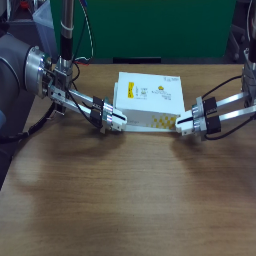}
    \caption{\textit{Push}: global image}
    \label{fig:push_camera}
  \end{subfigure}%
  \captionsetup{justification=centering}
  \caption{Sample of \textit{Set} and \textit{Push}. In $\textit{Push}$, the robot can infer the action using the foveated image because it includes both end-effectors.}
  \label{fig:sample_box}
 \end{figure}
 
Fig. \ref{fig:apple_orange_attention_weights} presents the \hl{normalized attention}, on the validation sets of \textit{Pick}, divided into the subtasks of apple-picking (using the left arm) and orange-picking (using the right arm).
This shows that when using the left arm (apple-picking), the Transformer architecture pays more attention to the states of the left arm, and vice versa. \hl{Then, the time series of the normalized attention on left/right arm was visualized for this task} (Fig. \ref{fig:apple_orange_temporal_attention}). \hl{Because the length of each episode is different, we interpolated the sequence of attention into a length of 100. The result shows the attention to the left arm decreases as subtask shifts from picking of apple (left) to orange (right), while the attention to the right arm increases.}

In the same way, we visualized the \hl{normalized attention} for each type of sensory information on \textit{BoxPush}. Fig. \ref{fig:box_attention_weights} visualizes the \hl{attention} on \textit{BoxPush}, where each episode into \textit{Set}, the behavior of placing the arms behind the box ($\sim6.0$s in Fig. {\ref{fig:box_push}}), and \textit{Push}, in which the robot actually pushes the box into the goal ($6.0$s$\sim$ in Fig. \ref{fig:box_push}). 
In the \textit{Set} subtask, the Transformer attended more to the left arm and right arm states than the image embedding. In contrast, in the \textit{Push} subtask, it attended to the image embedding because the foveated image alone is sufficient information for moving the box to the goal position (Fig. \ref{fig:sample_box}).

\section{Discussion}
We proposed the Transformer-based self-attention mechanism for deep imitation learning on real robot dual-arm manipulation tasks. \hl{Because the self-attention mechanism masks out sensory input which is not related to the current task, this suppresses distractions on sensory input.}
The experiments demonstrated that the proposed Transformer-based method substantially improves uncoordinated (\textit{Pick}), goal-coordinated (\textit{KnotTying}), and bimanual (\textit{BoxPush, ChangeHands}) dual-arm manipulation performance over the baseline without the Transformer. The analysis on the attention rollout revealed that the Transformer can attend to the appropriate sensory input.

Our Transformer-based deep imitation learning architecture is not specialized for dual arms but instead can be expanded to more complicated robots such as multi-arm robots or humanoid robots by concatenating more sensory information into the state representation. 
In the same way, it would be interesting to investigate whether additional sensory information such as tactile or sound can be integrated using our proposed method, which remains as future work.

In our experiment, closed-chain bimanual manipulation tasks such as moving heavy objects using both arms were not tested. In our teleoperation setup without force feedback, the counterforce from the object is not transferred back to the human, causing failure while teleoperating the closed-chain manipulation tasks. To solve this issue in the future, a bilateral system with force feedback may be required to enable a human to correctly control the system.
    

\bibliographystyle{IEEEtran}
\bibliography{IEEEfull}

\end{document}